\documentclass[conference]{IEEEtran}
\usepackage[utf8]{inputenc}
\IEEEoverridecommandlockouts

\usepackage{cite}

\usepackage{amsmath,amssymb,amsfonts}
\usepackage{algorithmic}
\usepackage{graphicx}
\usepackage{textcomp}
\usepackage{xcolor}
\usepackage{hyperref}
\usepackage{physics}
\usepackage{float}
\usepackage{inputenc}
\usepackage{bm}
\usepackage{adjustbox}
\usepackage{fancyhdr}

\usepackage{tikz}
\usetikzlibrary{arrows.meta,positioning,fit,backgrounds,shapes.geometric}

\usepackage{circuitikz}

\def\BibTeX{{\rm B\kern-.05em{\sc i\kern-.025em b}\kern-.08em
    T\kern-.1667em\lower.7ex\hbox{E}\kern-.125emX}}
\begin{document}

\title{Medical Imaging Classification with Cold-Atom Reservoir Computing using Auto-Encoders and Surrogate-Driven Training\\
}


\author{\IEEEauthorblockN{
    Nuno Batista$^{1}$,
    Ana Morgado$^{1}$,
    Oscar Ferraz$^{1}$,
    Sagar Silva Pratapsi$^{3}$,
    Jorge Lobo$^{2}$, and
    Gabriel Falcao$^{1}$}
    \\ 
    $^{1}$Instituto de Telecomunicações, 
   Dept. of Electrical and Computer Engineering,
   University of Coimbra, Portugal \\ 
    $^{2}$ ISR - Institute of Systems and Robotics,
   Dept. of Electrical and Computer Engineering,\\
   University of Coimbra, Portugal \\
    $^{3}$CFisUC, Department of Physics, University of Coimbra, Portugal \\
    Email: {\{nuno.batista, ana.morgado, oscar.ferraz, gff\}@co.it.pt}, sagar.pratapsi@uc.pt, jlobo@isr.uc.pt

\thanks{This work was supported by FCT - Fundação para a Ciência e Tecnologia, I.P. by project reference UID/50008/2023 IT, UIDB/04564/2020, UIDP/04564/2020, 2022.06780.PTDC, and 2023.14860.PEX (Q-Bet) with DOI identifiers 10.54499/UID/50008/2023, 10.54499/UIDB/04564/2020, 10.54499/UIDP/04564/2020, 10.54499/2022.06780.PTDC, and 10.54499/2023.14860.PEX, respectively. It was also funded by the Open Quantum Institute (OQI), and conducted in collaboration with Centro de Estudos Sociais (CES), University of Coimbra (UC), and Instituto de Telecomunicações (IT). OQI itself is an initiative hosted by CERN, born at GESDA, supported by UBS).(Corresponding author: Nuno Batista)}}


\maketitle

\thispagestyle{plain}
\fancypagestyle{plain}{
  \fancyhf{}
  \fancyhead[C]{\footnotesize Accepted to the 2025 IEEE International Conference on Quantum AI (IEEE QAI)}
  \renewcommand{\headrulewidth}{0pt} 
}

\begin{tikzpicture}[remember picture, overlay]
\node[anchor=south, yshift=10pt] at (current page.south) {
  \begin{minipage}{\textwidth}
  \centering \scriptsize
  \copyright 2025 IEEE. Personal use of this material is permitted. Permission from IEEE must be obtained for all other uses, in any current or future media, including reprinting/republishing this material for advertising or promotional purposes, creating new collective works, for resale or redistribution to servers or lists, or reuse of any copyrighted component of this work in other works.
  \end{minipage}
};
\end{tikzpicture}


\begin{abstract}
We introduce a hybrid quantum-classical pipeline, based on neutral-atom reservoir computing, for medical image classification, focusing on the binary classification task of polyp detection. To deal effectively with the high dimensionality, we integrate a guided auto-encoder. This pipeline learns compact and discriminative representations of image data that are also well-suited for quantum reservoir computing. A key challenge in such systems is the non-differentiable nature of quantum measurements, which creates a `gradient barrier' for standard training. We overcome this barrier by incorporating a differentiable surrogate model that emulates the quantum layer, enabling end-to-end backpropagation through the entire system. This guided training process is jointly optimized for classification accuracy and for faithful image recovery from the auto-encoder. The learned latent representations are encoded as pulse detuning parameters within a Rydberg Hamiltonian, and quantum embeddings are subsequently obtained through expectation values. These embeddings are then passed to a linear classifier. Our simulations show that this method outperforms some traditional approaches that use PCA or unguided autoencoders. We also conduct ablation studies to assess the impact of various quantum and training parameters, demonstrating the robustness and flexibility of our proposed pipeline for real-world medical imaging applications, even in the current NISQ era.
\end{abstract}

\begin{IEEEkeywords}

Reservoir Computing, Quantum-Guided Autoencoding, Neutral Atoms, Dimensionality Reduction, Quantum Machine Learning, Quantum Medical Image Classification, Quantum Surrogate Models, Quantum Computing, QML for NISQ

\end{IEEEkeywords}

\section{Introduction}

Advances in medical imaging have significantly improved disease diagnosis and treatment planning. For conditions like colorectal cancer, early detection of polyps through colonoscopy image analysis is critical for reducing mortality~\cite{estevaGuideDeepLearning2019a}. Deep learning techniques, especially autoencoders, are widely used to extract compressed, informative features from high-dimensional images for classification~\cite{bengioLearningDeepArchitecturesa}. However, classical neural networks may struggle to capture intricate correlations in complex medical data~\cite{meiFrameworkProcessingLargescale2024}.

Quantum computing offers novel opportunities for machine learning, particularly through quantum reservoir computing (QRC), where a physical quantum system processes classical inputs into high-dimensional nonlinear embeddings~\cite{tanakaRecentAdvancesPhysical2019,fujiiHarnessingDisorderedQuantum2017}. Recent works show that analog quantum systems, such as neutral-atom platforms, can serve as untrained reservoirs with rich dynamics for temporal and pattern recognition tasks~\cite{domingoOptimalQuantumReservoir2022,kornjavcaLargescaleQuantumReservoir2024}. 

In this work, we propose a quantum-guided autoencoder architecture that integrates a classical image encoder with a neutral-atom quantum reservoir.

In hybrid approaches, a classical encoder compresses image data, and a quantum reservoir expands the encoded features into a higher-dimensional Hilbert space, potentially boosting classification performance, as these expanded features might capture high-order correlations and nonlinearities that classical methods cannot tractably represent.

A major challenge in such hybrid quantum-classical models is the non-differentiability of quantum measurements, which obstructs gradient-based optimization. Additionally, tuning quantum parameters can suffer from barren plateaus, where gradients vanish in high-dimensional Hilbert spaces~\cite{mccleanBarrenPlateausQuantum2018}. To address this, we introduce a classical neural surrogate that emulates the quantum reservoir's input-output behavior. This surrogate enables end-to-end training via backpropagation, while the quantum system remains fixed and non-trainable. It is only used to train the surrogate model and to create embeddings for downstream classifiers.

Critically, the physical reservoir’s exponentially large state‐space (and associated many-body correlations) cannot be classically emulated, hence the surrogate is merely a differentiable proxy during training, not a replacement of the real quantum dynamics at inference.



Our results illustrate the viability of QRC for real-world medical tasks and offer a scalable path to hybrid quantum-classical learning, even in the noisy intermediate-scale quantum (NISQ) era.

\section{Background and Related Work}
This section provides an overview of the foundational concepts and prior research relevant to this work.
\subsection{Principles of Reservoir Computing}
Reservoir computing is a computational framework derived from recurrent neural networks (RNNs). It involves a fixed, high-dimensional dynamical system (the 
reservoir) that projects input data into a rich feature space. Only the output layer is trained, simplifying the learning process and reducing computational overhead. This approach is particularly effective for time-series prediction and pattern recognition tasks.

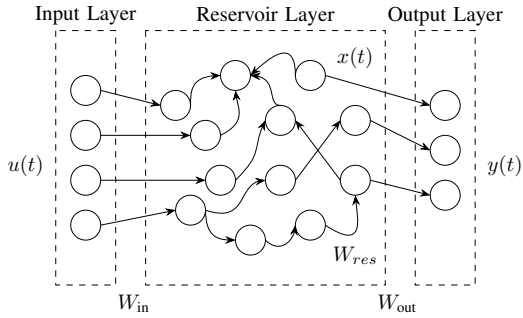
\begin{figure}[!h]
    \centering
    \resizebox{0.8\columnwidth}{!}{
        \begin{circuitikz}
            \tikzstyle{every node}=[font=\normalsize]
            \draw(6,-1.5) circle (0.25cm);
            \draw(6,-2.25) circle (0.25cm);
            \draw(6,-3) circle (0.25cm);
            \draw(6,-0.75) circle (0.25cm);
            \draw(7.5,-1) circle (0.25cm);
            \draw(8,-1.5) circle (0.25cm);
            \draw(8.25,-2.25) circle (0.25cm);
            \draw(7.75,-2.75) circle (0.25cm);
            \draw(8.5,-0.5) circle (0.25cm);
            \draw(9.25,-2.25) circle (0.25cm);
            \draw(9.75,-0.5) circle (0.25cm);
            \draw(9.25,-1.25) circle (0.25cm);
            \draw(9.75,-3) circle (0.25cm);
            \draw(8.75,-3.25) circle (0.25cm);
            \draw(10.5,-2.25) circle (0.25cm);
            \draw(10.5,-1.25) circle (0.25cm);
            \draw[->, >=Stealth] (8,-2.75) .. controls (8,-3.25) and (8.25,-3.25) .. (8.5,-3.25) ;
            \draw[->, >=Stealth] (8,-2.75) .. controls (8.75,-3) and (8.5,-2.25) .. (9,-2.25) ;
            \draw[->, >=Stealth] (9,-3.25) .. controls (9.25,-3.5) and (9.5,-3.25) .. (9.5,-3) ;
            \draw[->, >=Stealth] (10,-3) .. controls (10.75,-3.5) and (10.5,-2.75) .. (10.5,-2.5) ;
            \draw[->, >=Stealth] (9.5,-2.25) -- (10.25,-1.25);
            \draw[->, >=Stealth] (10.25,-2.25) -- (9.5,-1.25);
            \draw[->, >=Stealth] (8.5,-2.25) .. controls (9,-1.75) and (8.75,-1.75) .. (9,-1.25) ;
            \draw[->, >=Stealth] (9.25,-1) .. controls (9,-1) and (9.25,-0.5) .. (8.75,-0.5) ;
            \draw[->, >=Stealth] (7.75,-1) .. controls (7.75,-0.75) and (7.75,-0.5) .. (8.25,-0.5) ;
            \draw[->, >=Stealth] (8.25,-1.5) .. controls (8.75,-1.25) and (8.5,-1) .. (8.5,-0.75) ;
            \draw[->, >=Stealth] (9.5,-0.5) .. controls (9.5,0) and (9.25,0) .. (8.75,-0.5) ;
            \draw(12,-1) circle (0.25cm);
            \draw(12,-1.75) circle (0.25cm);
            \draw(12,-2.5) circle (0.25cm);
            \draw[->, >=Stealth] (10,-0.5) -- (11.75,-1);
            \draw[->, >=Stealth] (10.75,-1.25) -- (11.75,-1.75);
            \draw[->, >=Stealth] (10.75,-2.25) -- (11.75,-2.5);
            \draw[ dashed] (5.5,0.25) rectangle  (6.5,-4);
            \draw[ dashed] (7,0.25) rectangle  (11,-4);
            \draw[ dashed] (11.5,0.25) rectangle  (12.5,-4);
            \draw[->, >=Stealth] (6.25,-0.75) -- (7.25,-1);
            \draw[->, >=Stealth] (6.25,-1.5) -- (7.75,-1.5);
            \draw[->, >=Stealth] (6.25,-2.25) -- (8,-2.25);
            \draw[->, >=Stealth] (6.25,-3) -- (7.5,-2.75);
            \node[font=\normalsize] at (6,0.5) {Input Layer};
            \node[font=\normalsize] at (9,0.5) {Reservoir Layer};
            \node[font=\normalsize] at (12,0.5) {Output Layer};
            \node[font=\normalsize] at (5,-2) {$u(t)$};
            \node[font=\normalsize] at (10.5,-3.5) {$W_{res}$};
            \node[font=\normalsize] at (13,-2) {$y(t)$};
            \node[font=\normalsize] at (6.8,-4.3) {$W_\text{in}$};
            \node[font=\normalsize] at (10.5,-0.2) {$x(t)$};
            \node[font=\normalsize] at (11.2,-4.3) {$W_\text{out}$};
        \end{circuitikz}
    }%

\caption{
    Reservoir computing architecture showing input nodes, 
    a recurrent reservoir network with internal dynamics, 
    and an output layer.
}
\label{fig:reservoir_architecture}
\end{figure}

Mathematically, let \( u(t) \in \mathbb{R}^m \) be the input at time \( t \), with
\( x(t) \in \mathbb{R}^n \) being the reservoir state, and
\( y(t) \in \mathbb{R}^k \) the output. The reservoir dynamics
and output are given by
\begin{align}
    x(t) &= f(W_{in} u(t) + W_{res} x(t-1)) \\
    y(t) &= W_{out} x(t),
\end{align}
%
%
where \( f \) is a nonlinear activation function,
\( W_{in} \) and \( W_{res} \) are fixed input and reservoir weight matrices, 
and \( W_{out} \) is the trained output weight matrix.
A diagram of a typical reservoir computing architecture is shown in Fig.~\ref{fig:reservoir_architecture}.

\subsection{Quantum Reservoir Computing with neutral atoms}

Quantum Reservoir Computing (QRC) extends the reservoir 
computing paradigm into the quantum domain. 
QRC aims to enhance computational capabilities by accessing larger state spaces and different types of non-linearities. 
Notably, large-scale experiments utilizing neutral-atom analog quantum computers have demonstrated the scalability and effectiveness of QRC in various machine learning applications~\cite{kornjavcaLargescaleQuantumReservoir2024}.

In QRC, classical input data \( u(t) \) is encoded into quantum states \( |\psi(t)\rangle \),
which evolve under a fixed Hamiltonian \( H \):
\begin{equation}
    \ket{\psi(t+\Delta t)}
    = U \ket{\psi(t)}
    = e^{-iH\Delta t} \ket{\psi(t)},
\end{equation}
where \( U \) is the unitary evolution operator and \(\Delta t\) is the difference between consecutive timestamps. Measurements of observables \( \hat{O} \) yield the outputs
\begin{equation}
    y(t) = \langle \psi(t) | \hat{O} | \psi(t) \rangle.
\end{equation}
The output weights are trained classically, while the quantum reservoir remains fixed.

Neutral atom platforms, particularly those utilizing Rydberg states, have emerged as promising candidates for implementing QRC due to their scalability and controllable interactions.

In the work by M. Kornjača et al.~\cite{kornjavcaLargescaleQuantumReservoir2024}, a large-scale, gradient-free QRC algorithm was developed and experimentally implemented on a neutral-atom analog quantum computer. This system achieved competitive performance across various machine learning tasks, including classification and time-series prediction, demonstrating effective learning with increasing system sizes up to $108$ qubits.


The dynamics of such laser-driven neutral atoms, working in the Rydberg state regime, is given by the Hamiltonian~\cite{kornjavcaLargescaleQuantumReservoir2024}

\begin{align}
    H(t)
    &=
    \frac{\Omega(t)}{2}
    \sum_j
    \left(
        \ketbra{g_j}{r_j}
        + \ketbra{r_j}{g_j}
    \right) \nonumber \\
    & \quad + \sum_{j<k}
    V_{jk} n_j n_k
    - \sum_j
    \left[
        \Delta_{\mathrm{g}}(t)
        + \alpha_j \Delta_{\mathrm{l}}(t)
    \right] n_j,
\label{eq:rydberg_hamiltonian}
\end{align}
where $\Omega(t)$ is the global Rabi drive amplitude between a 
ground state $\ket{g_j}$ and a highly-excited Rydberg state 
$\ket{r_j}$ of an atom $j$,
$n_j = \ketbra{r_j}{r_j}$, and 
\( V_{jk} = C_6/\|r_j - r_k\|^6 \) describes the van der Waals interactions
between atoms, and the detuning is split into a global term 
\( \Delta_{\mathrm{g}}(t) \)
and a site-dependent term \( \Delta_{\mathrm{l}}(t) \), with site 
modulation \( \alpha_j \in [0, 1] \).

By initializing the system in a specific state and allowing 
it to evolve under this Hamiltonian, the resulting quantum 
state encodes information about the input data. Measurements 
of observables on this state yield outputs that can be used 
for tasks such as classification or prediction, with only the 
final readout layer requiring training.

\subsection{Dimensionality Reduction for Image Data}
Dimensionality reduction is a crucial preprocessing step in 
machine learning and data analysis, aiming to reduce the 
number of input variables in a dataset while preserving as 
much information as possible. This process enhances 
computational efficiency and facilitates data visualization.

\subsubsection{Principal Component Analysis}
Principal Component Analysis (PCA)~\cite{shlensTutorialPrincipalComponent2014} is a linear dimensionality 
reduction technique that transforms a set of correlated 
variables into a set of uncorrelated variables called 
principal components. The goal is to capture the maximum 
variance in the data with the fewest number of components.

PCA is effective for datasets where the principal components 
align with the directions of maximum variance, but it may not 
capture complex, nonlinear relationships in the data~\cite{jolliffePrincipalComponentAnalysis2016}.

\subsubsection{Autoencoder Architectures}
Autoencoders are a class of artificial neural networks 
designed to learn efficient codings of input data in an 
unsupervised manner. They consist of two main parts: an 
encoder that compresses the input into a latent-space 
representation, and a decoder that reconstructs the input 
from this representation.

Given an input \( \bm x \in \mathbb{R}^d \), the encoder maps 
\( \bm x \) to a latent representation \( \bm z \in \mathbb{R}^k \) 
(where \( k < d \)):

\begin{equation}
    \bm z = \varepsilon_{\bm\omega}(\bm x).
\end{equation}
The decoder then reconstructs the input $\bm{\hat{x}} \in \mathbb{R}^d$:
\begin{equation}
    \bm{\hat{x}} = \mathcal{D}_{\bm \rho} (\bm z),
\end{equation}
where we follow the notation from~\cite{belisGuidedQuantumCompression2024}.
The network is trained to minimize the reconstruction loss:
\begin{equation}
    \mathcal{L}(\bm{x}, \bm{\hat{x}}) = \| \bm{x} - \bm{\hat{x}} \|^2.
\end{equation}
This architecture of an autoencoder is illustrated in Fig. 2 a). 

Autoencoders can capture complex, nonlinear 
relationships in the data, making them suitable for tasks 
like image compression, denoising, and anomaly 
detection~\cite{hintonReducingDimensionalityData2006, estevaGuideDeepLearning2019a}.

\subsubsection{Quantum-Guided Autoencoding}
Quantum-Guided Autoencoders integrate quantum computing principles into the autoencoder framework to leverage quantum advantages in processing and representing data. These models aim to perform dimensionality reduction and classification within a single architecture, enhancing performance on complex datasets.

In the Quantum-Guided Autoencoder (QGA) model, a classical encoder first reduced the dimensionality of the input data. The compressed data is then processed by a parametrized quantum circuit, which acts as the decoder and classifier. The quantum circuit transforms the input state \( |\psi_{\text{in}}\rangle \) into an output state \( |\psi_{\text{out}}\rangle \) using a unitary operation:
\begin{equation}
    |\psi_{\text{out}}\rangle = U(\bm\theta) |\psi_{\text{in}}\rangle.
\end{equation}
The parameters \( \bm\theta \) are optimized to minimize a loss function that combines reconstruction error and classification accuracy. Measurements on \( |\psi_{\text{out}}\rangle \) yield the final classification result. 

This approach has demonstrated an improved performance over traditional methods in tasks such as identifying the Higgs boson in particle collision data, showcasing the potential of quantum-guided models in handling high-dimensional, complex datasets~\cite{belisGuidedQuantumCompression2024}.

\section{Methodology}
\subsection{Proposed System Architecture}
Our proposed pipeline is a Quantum-Guided Autoencoder with a 
Reservoir Surrogate (QGARS). It is an hybrid architecture that 
synergizes classical autoencoding techniques with quantum reservoir processing to achieve efficient learning of features. These are expected to better match with the reservoir computing layer, whose embeddings are used to perform a classification task.

The system is designed to overcome the inherent non-differentiability of quantum operations, which traditionally impede gradient-based training methods.
The overall architecture is illustrated in Fig. 2 and the \textbf{architectural components} are the following:
\begin{itemize}
    \item \textbf{Classical Autoencoder:}
        \begin{itemize}
            \item \textbf{Encoder:}
            transforms high-dimensional input data \(\bm{x} \) into 
            a lower-dimensional latent representation \(\bm{z}\).
            \item \textbf{Decoder:}
            The decoder network \(\mathcal{D}_\rho(\bm{z})\) reconstructs the input data from the latent representation, producing \(\bm{\hat{x}}\).
        \end{itemize}

    \item \textbf{Quantum Reservoir Layer:}
        \begin{itemize}
            \item \textbf{Parameter Mapping:}
            \(g(\bm{z})\) maps the latent representation \(\bm{z}\) to the local detuning frequencies
            \(\Delta_{\mathrm{i}}\) of the Rydberg atoms.
            \item \textbf{Quantum Evolution:}
            Evolves the system under a Hamiltonian \(H(\bm{\Delta)}\), resulting in a quantum 
            state \(|\psi(\bm{\Delta})\rangle\).
            \item \textbf{Measurement:}
            Performs measurements on the quantum state to obtain embeddings \(\bm{q} \in \mathbb{R}^k\), which serve as training targets for the surrogate model or as 
            inputs for the classification task later on.
        \end{itemize}

    \item \textbf{Surrogate Model:}
    A differentiable classical neural network \( f_{\text{sur}}(\bm{z}; \bm{\theta}_{\text{sur}}) \),
    where \(\bm{\theta}_{\text{sur}}\) are parameters tuned to approximate the 
    mapping from \(\bm{z}\) to \(\bm{q}\), effectively emulating
    the quantum layer's behaviour.

    \item \textbf{Linear Classifier:}
    A classical neural network that maps the quantum embeddings \(\bm{q}\) to 
    predicted class labels \(\bm{\hat{y}}\).

\end{itemize}
The following sections give a more detailed explanation of each component.

\begin{figure*}[!t]
    \centering
    \resizebox{0.9\textwidth}{!}{
        \includegraphics{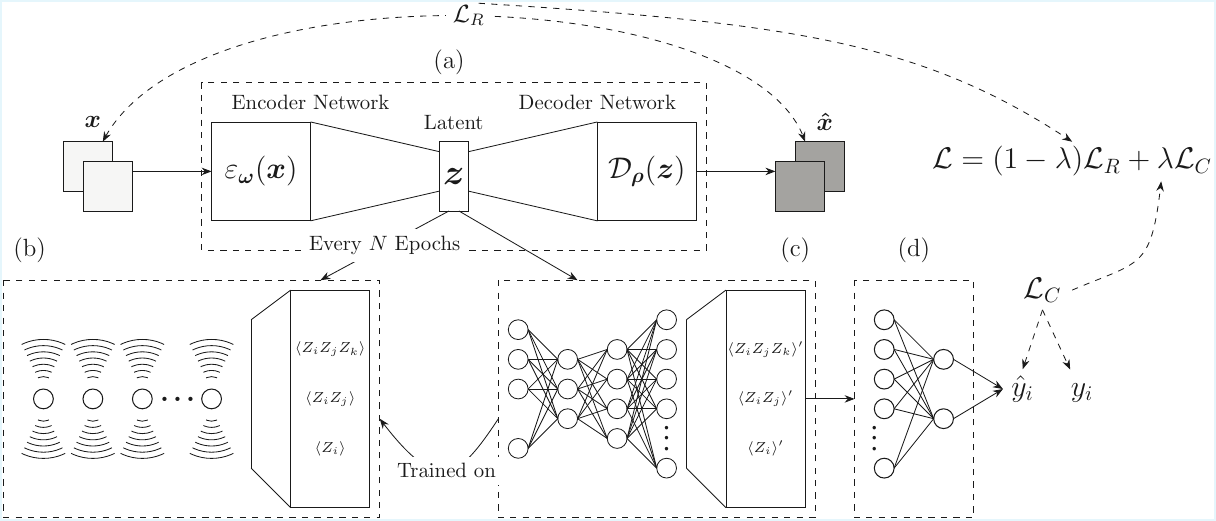}
    }
    \label{fig:qgars}
    \caption{
    Overview of the QGARS pipeline. (a) Classical autoencoder: $\varepsilon_{\bm\omega}(\bm{x})$ is the encoder network that maps input $\bm{x}$ to latent code $\bm{z}$, and $\mathcal{D}_{\bm\rho}(\bm{z})$ is the decoder network reconstructing $\hat{\bm{x}}$. (b) Rydberg atom quantum reservoir: the latent code $\bm{z}$ is mapped to detuning parameters $\Delta_i$, the system evolves under the Hamiltonian in Eq.~\eqref{eq:rydberg_hamiltonian}, and measurements of observables $\langle Z_i\rangle,\langle Z_iZ_j\rangle,\langle Z_iZ_jZ_k\rangle$ yield quantum embeddings. (c) Surrogate model: a feedforward neural network $f_{\mathrm{sur}}(\bm{z};\bm\theta_{\mathrm{sur}})$ trained every $N$ epochs to approximate the quantum embeddings. (d) Linear classifier: maps surrogate outputs $\bm{q}'$ to class predictions, closing the loop with total loss $\mathcal{L}=(1-\lambda)\mathcal{L}_R+\lambda\mathcal{L}_C$.
    }
\end{figure*}

\subsection{Guided Autoencoder}
The Guided Autoencoder combines classical with quantum processing to leverage the strengths of both paradigms.

Ideally, the latent space representation created by the encoder \(\varepsilon_\omega(\bm{x})\)
would be passed through two sections of the model:
\begin{itemize}
    \item \textbf{The Decoder Network \(\mathcal{D}_\rho(\bm{z})\),}
    which tries to reconstruct the original input data.
    \item \textbf{The Quantum Reservoir,}
    which extracts highly dimensional quantum embeddings and passes them to a linear classifier to perform a classification task and map the embeddings to the image labels.
\end{itemize}

However, some problems emerge when using a real quantum reservoir directly during the autoencoder training stage, which will be discussed in a later section.

\subsubsection{Loss Function Design}
The QGA is trained to minimize a composite loss 
function that balances reconstruction fidelity and 
classification accuracy:

\begin{equation}
    \mathcal{L} = (1-\lambda) \cdot \mathcal{L}_R + \lambda \cdot \mathcal{L}_{C},
    \label{eq:composed_loss_function}
\end{equation}
where:
\begin{itemize}
    \item \( \mathcal{L}_R = \| \bm{x} - \hat{\bm{x}} \|^2 \) is the reconstruction loss, measuring the mean squared error between the input \( \bm{x} \) and its reconstruction \( \bm{\hat{x}} \).
    \item \( \mathcal{L}_C = -\sum_i y_i \log(\hat{y}_i) \) is the classification loss, computed as the cross-entropy between the true labels \( y_i \) and the predicted probabilities \( \hat{y}_i \), which are obtained through a linear mapping of the 
    approximated embeddings \(\bm{q}'\).
    \item \( 0 < \lambda < 1 \) is a hyperparameter that controls the trade-off between reconstruction and classification objectives.
\end{itemize}    
\subsection{The Gradient Barrier Problem}
Hybrid quantum-classical models, such as the Quantum-Guided Autoencoder (QGAE), aim to leverage quantum computational advantages within classical machine learning frameworks. However, a significant challenge arises due to the non-differentiable nature of quantum 
operations, we refer to this as the `gradient barrier'.

In classical neural networks, training relies on backpropagation, 
which requires the computation of gradients through all components 
of the network. In hybrid models, the inclusion of quantum 
layers introduces operations that are inherently non-differentiable:

\begin{itemize}
    \item \textbf{Quantum State Preparation}: 
    Encoding classical data into 
    quantum states often involves operations 
    that are not smoothly differentiable with 
    respect to the input data.
    
    \item \textbf{Quantum Measurements}:
    Observing quantum states collapses them, 
    introducing stochasticity and breaking the 
    deterministic gradient flow required 
    for backpropagation.
\end{itemize}

These aspects hinder the direct application of 
gradient-based optimization methods across the 
quantum-classical boundary, obstructing end-to-end 
training of hybrid models. The surrogate model 
solution, proposed in the next section~\ref{surrogate_modeling_for_quantum_layers}, addresses 
this problem.

\subsection{Surrogate Modeling for Quantum Layers} \label{surrogate_modeling_for_quantum_layers}
To circumvent the gradient barrier, surrogate models 
have been proposed. These are classical, differentiable 
models trained to approximate the input-output behavior 
of quantum circuits. By replacing the non-differentiable 
quantum components with their surrogate counterparts during 
training, gradients can be propagated through the entire network, 
enabling end-to-end optimization. 

Recent studies have demonstrated the efficacy of surrogate models in mitigating the effects of barren plateaus and facilitating the training of hybrid quantum-classical models. These approaches leverage
classical optimization techniques while still capturing quantum computational advantages~\cite{xieQuantumSurrogateDrivenImage2025a}.

\subsubsection{Surrogate Model Architecture}
In our pipeline, the surrogate model \( f_{\text{sur}}(\bm{z}; \bm\theta_{\text{sur}}) \) is implemented as a feedforward neural network designed to learn the mapping $\bm q´$ from the autoencoder's latent space to the quantum embeddings $\bm q$ produced by the quantum reservoir:
\begin{equation}
    f_{\text{sur}}(\bm{z}; \bm\theta_{\text{sur}}) = \bm{q}' \approx \bm{q} =  \langle \psi(g(\bm{z}))| \hat{O} |\psi(g(\bm{z}))\rangle
\end{equation}
where \(\hat{O}\) represents the observables whose expectation values are to be determined.

Its architecture comprises multiple fully connected layers with nonlinear activation functions, facilitating the approximation of the complex, non-linear transformations inherent in quantum dynamics.


\subsubsection{Training Procedure}
\begin{itemize}
    \item \textbf{Data Input}: 
    Feed a batch of latent data \( \bm{z} \) from the autoencoder 
    into the real quantum reservoir, recording the resulting quantum 
    embeddings.

    \item \textbf{Surrogate Training}:
    Train the surrogate model  by minimizing the loss function:
    \begin{equation}
        \mathcal{L}_{\text{sur}} = \| \bm{q}' - \bm{q} \|^2,
    \end{equation}

    \item \textbf{Regular Updates}:
    Periodically update the surrogate model during training
    to ensure it accurately reflects the quantum reservoir's
    behavior. 
\end{itemize}

\subsubsection{Gradient Flow Through Surrogate Models}
By integrating the surrogate model into the training pipeline, we establish a continuous computational graph from the output layer back to the input layer, enabling the use of gradient-based optimization techniques. The surrogate model serves as a differentiable proxy for the quantum reservoir, allowing the classification loss to influence the autoencoder's parameters effectively.


\subsection{Rydberg Hamiltonian and Quantum Dynamics}
While a fixed classical reservoir can nonlinearly mix inputs, its capacity is polynomially bounded. In contrast, our neutral-atom reservoir leverages many-body quantum dynamics to encode data in a space whose dimension grows exponentially with atom number, unlocking patterns that classical reservoirs would need unreasonably complex architectures to approximate. This means that the quantum reservoir is a crucial part of our pipeline and that the surrogate is just an approximation to aid in the autoencoder training process.


\subsubsection{Data Encoding Schemes}
The input data is encoded into the quantum system by mapping
the latent representations from the autoencoder to the
detuning parameters of the Rydberg Hamiltonian.
This encoding process involves adjusting the detuning
parameters \( \Delta_{\mathrm{l}}(t) \) for each atom in the chain,
allowing the system to represent the input data in a way that
is compatible with the quantum dynamics of the reservoir.

\subsubsection{Quantum Readout Methods and Embedding Dimensionality}
The quantum reservoir's output is obtained by probing, 
in successive time steps, the state of the atoms 
after the systems evolution up to a certain time \( t \).
The readout is performed by measuring specific observables,
such as $\langle Z_i \rangle$, $\langle Z_i \, Z_j\rangle$, and $\langle Z_i \, Z_j \, Z_k \rangle$, which measure respectively one-, two- and three-qubit correlations. 

This means that the dimensionality of the QRC embeddings generated in our pipeline is a direct consequence of the number of qubits ($N$) employed in the neutral-atom chain, the specific quantum mechanical observables measured, and the number of discrete time steps ($T$) over which these measurements are collected every \(\Delta t\).

If we measure all of the three proposed observables, the total count of the values measured at a single time step $O_N$ is the sum of the counts for each type of observable:
\begin{equation}
    \begin{aligned}
    O_N &= N + \binom{N}{2} + \binom{N}{3}.
    \end{aligned}
    \label{eq:embeddings_per_timestep}
\end{equation}
This set of $O_N$ measurements is repeated at $T$ successive time steps during the evolution of the quantum reservoir. Consequently, the total dimensionality, $\lvert \bm q \rvert$, of the final QRC embedding vector is:
\begin{equation}
    \lvert \bm q \rvert = T O_N.
    \label{eq:total_embedding_dimension}
\end{equation}
This formulation shows that the embedding dimension scales polynomially with the number of qubits $N$ and the number of timesteps $T$ as $\mathcal{O}(TN^3)$ for the considered observables.

\section{Experimental Setup}
\subsection{Datasets}
We evaluate our proposed architecture on three datasets:
\begin{enumerate}
    \item \textbf{Synthetic Polyps:} 
    A synthetic dataset of polyp images generated to simulate 
    realistic medical imaging scenes.
    
    \item \textbf{CVC-ClinicDB:} 
    Real image patches extracted from the CVC-ClinicDB 
    dataset, a well-known benchmark for polyp detection. 
    
    \item \textbf{MNIST:} 
    A reduced version of the MNIST dataset, 
    containing only the digits 0 and 1, suitable 
    for binary classification tasks.
\end{enumerate}

\subsection{Feature Reduction Methods}
We implemented and compared three distinct feature reduction approaches:
\subsubsection{Principal Component Analysis (PCA)}
We extracted the top-k principal components 
(typically 4-12) from the flattened image data, 
which were then used as inputs to both classical and 
quantum-enhanced classifiers.

\subsubsection{Autoencoder}
We implemented a simple neural network-based autoencoder with a encoder network that compresses the input to a lower dimensional latent space and a decoder network that reconstructs the original input from the latent representation. To improve training stability and prevent overfitting, we also added batch normalization and dropout layers.

The autoencoder was trained to minimize reconstruction loss using the Adam optimizer with learning rates between $0.001$--$0.01$ and weight decay for regularization.

\subsubsection{Quantum-Guided Autoencoder}
We introduced a quantum-guided autoencoder that incorporates feedback from the quantum reservoir during training. This approach combines:

\begin{itemize}
    \item A standard autoencoder architecture.
    \item A linear classifier that maps quantum embeddings to class labels.
    \item A classification loss computed on quantum embeddings.
    \item A weighting parameter $\lambda$ to control the trade-off between reconstruction and classification objectives~\eqref{eq:composed_loss_function}.
    \item A differentiable surrogate model that approximates the quantum reservoir's behavior, enabling end-to-end backpropagation through the entire pipeline. The frequency of quantum updates is controlled by a hyperparameter, allowing for flexible integration of quantum dynamics into the training process.
\end{itemize}

\subsection{Quantum Reservoir Computing Layer}
Our quantum reservoir computing implementation uses a Rydberg atom simulator with the following parameterizable components of an atomic chain:

\begin{itemize}
    \item \textbf{Atom Chain Length (number of qubits):} 
    The number of atoms in the chain.
    
    \item \textbf{Rabi Frequency:} 
    The global Rabi drive amplitude, which influences the strength of interactions between atoms.
    
    \item \textbf{Lattice Spacing:} 
    The empty space between two consecutive atoms.
    
    \item \textbf{Measurement Strategy:} 
    The specific observables measured to obtain quantum embeddings.

    \item \textbf{Time steps:}
    The number of time steps where the reservoir will be probed.

    \item \textbf{Evolution Time:}
    The time we let the quantum reservoir evolve under the Hamiltonian~\eqref{eq:rydberg_hamiltonian}.
\end{itemize}

Unless explicitly stated, the results that we show in the next section~\ref{results_and_discussion} were obtained with the following hyperparameter values: $12$ qubits, $10 \mu m$ between atoms, \(\langle Z_i\rangle\), \(\langle Z_iZ_j\rangle,\) and \(\langle Z_iZ_jZ_k\rangle\) observables, Rabi Frequency $\Omega = 2\pi$~rad$/\mu s$, $T = 16$ time steps and evolved over $4.0 \mu s$ (which means that the system was probed every $\frac{4.0}{T} = \Delta t = 0.25 \mu s$.

We can use~\eqref{eq:total_embedding_dimension} to determine the dimension of our quantum embeddings \(\lvert \bm q \rvert \) when using the mentioned parameters:

\begin{equation}
    \lvert \bm q \rvert = \frac{16 \times 12^3 + 5 \times 16 \times 12}{6} = 4768
    \label{eq:experimental_embedding_dimension}
\end{equation}

\subsection{Linear Classifier for the QRC Embeddings} \label{linear_classifier_for_the_qrc_embeddings}
After the traning process of the QGARS model,
the following quantum embeddings are passed to a linear classifier.
The classifier is a simple feedforward neural network without hidden layers with the following architecture:
\begin{itemize}
    \item An input layer with the size of the embeddings.
    \item An output layer with two neurons.
    \item Cross-entropy loss for training, optimized using the Adam optimizer.
\end{itemize}
In total, this linear classifier has \(9538\) parameters when the reservoir contains $12$ qubits.

\subsection{Comparison Methods}
We benchmarked the prediction accuracy of our model against the following classical methods:
\begin{itemize}
    \item \textbf{PCA/Autoencoder + Linear Classifier:} 
    A linear classifier trained on the PCA-reduced features or the latent representations from the autoencoder. Essentially, this model is the same as the one used to classify the quantum embeddings described in section~\ref{linear_classifier_for_the_qrc_embeddings}, but it only receives the classically extracted features as input. 
    
    \item \textbf{PCA/Autoencoder + Neural Network:}
    A fully connected 4-layer neural network with two $100$-neuron hidden layers, trained on the features from the same reduction methods as the linear classifier. This neural network has $11602$ parameters when training on $12$ dimensional inputs.
\end{itemize}

We also benchmarked our QGARS encoding strategy
against the performance of the Linear Mapping of
QRC embeddings extracted from the latent representation 
of data using other feature reduction methods, such as
PCA and the classical autoencoder.

The accuracy results were measured with 2000 training images and 400 testing images over 5 different seeds. 


    
    
    

\subsection{Parameter Sweep Strategy}
To optimize the performance of our quantum-guided autoencoder, 
we conducted a systematic parameter sweep over the following hyperparameters:
\begin{itemize}
    \item \textbf{Guided Lambda Parameter (\( \lambda \))}: 
    To analyse the effect that the trade-off between reconstruction and classification objectives in the loss function~\eqref{eq:composed_loss_function} affects
    the QRC
    
    \item \textbf{Quantum Update Frequency:} 
    The frequency at which the quantum reservoir is updated during training. We experimented with update frequencies of 5, 7, 10 and 25 epochs.


    \item \textbf{Number of Features:}
    The number of features is the same as the number of qubits (atoms), 
    as each feature encoded in each atom's local detuning parameter.
    
\end{itemize}
The results from these experiments are discussed in the next section~\ref{results_and_discussion}.

\section{Results and Discussion} \label{results_and_discussion}

\subsection{Dimensionality Reduction Method Comparison}
We first compare classification accuracy of the linear classifier on the QRC embeddings across the different dimensionality reduction methods PCA, vanilla autoencoder, and our QGARS pipeline on each dataset. Table~\ref{tab:qrc_reduction_accuracy} summarizes mean accuracy and standard deviation over five runs, using 12 features/qubits. These results show that the QGARS pipeline is able to outperform traditional feature reduction methods commonly used as preprocessing steps on QRC pipelines.

\begin{table}[!b]
  \caption{Mean ± standard deviation of the classification accuracy (\%) QRC with different dimensionality reduction methods}
  \centering
  \label{tab:qrc_reduction_accuracy}
  \begin{tabular}{lccc}
    \hline
    Method         & Synthetic  & CVC-ClinicDB & MNIST(0/1) \\
    \hline
    PCA + QRC      &  79.20 ± 2.53  &  84.95 ± 2.31   &  97.70 ± 0.97 \\
    AE + QRC       &  77.50 ± 2.70  & 73.75 ± 1.90   &  87.05 ± 2.09 \\
    \textbf{QGARS + QRC}    &  \textbf{89.45 ± 2.06} &  \textbf{88.90 ± 1.65} &  \textbf{97.00 ± 2.37} \\
    \hline
  \end{tabular}
\end{table}

\begin{figure}[!t]
    \centering
    \includegraphics[width=0.8\linewidth]{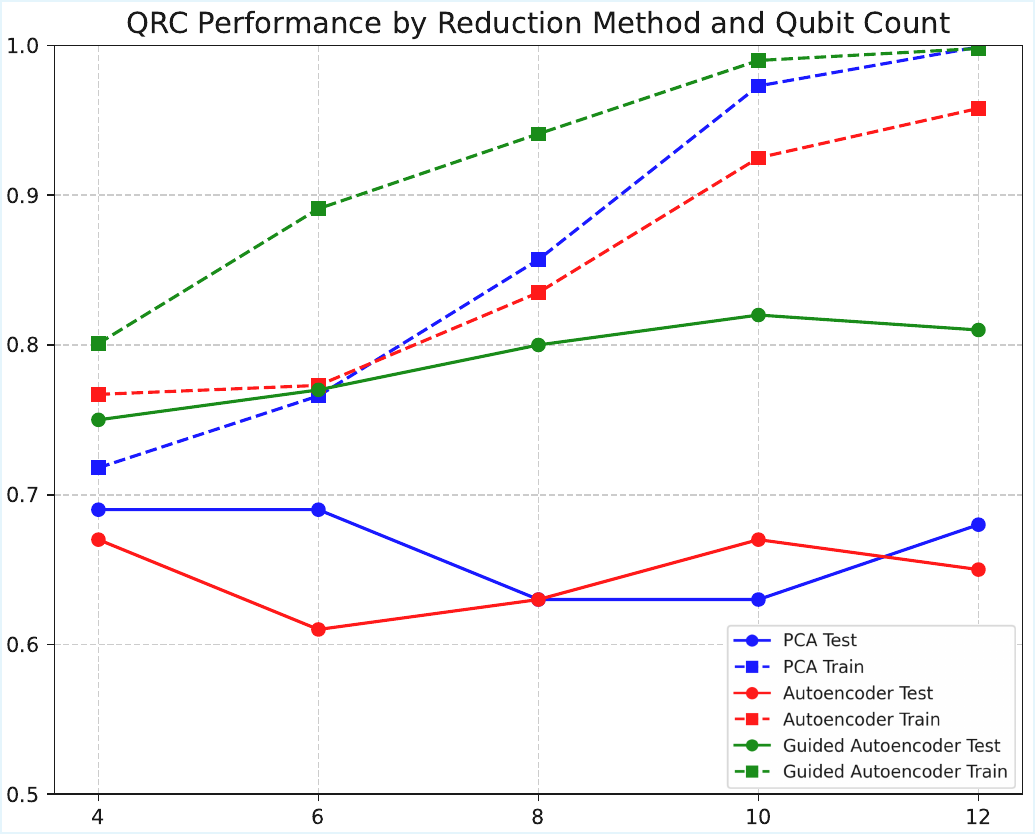} 
    \caption{Accuracy of the different dimensionality reduction methods for QRC as a function of the number of qubits with the synthetic polyp dataset. (green) QGARS, (blue) PCA and (red) Autoencoder. The dashed lines represent the training accuracy, while the whole lines are the test accuracies.}
    \label{fig:n_qubits_accuracy}
\end{figure}

\begin{figure*}[!ht]
  \centering
  \includegraphics[width=0.32\textwidth]{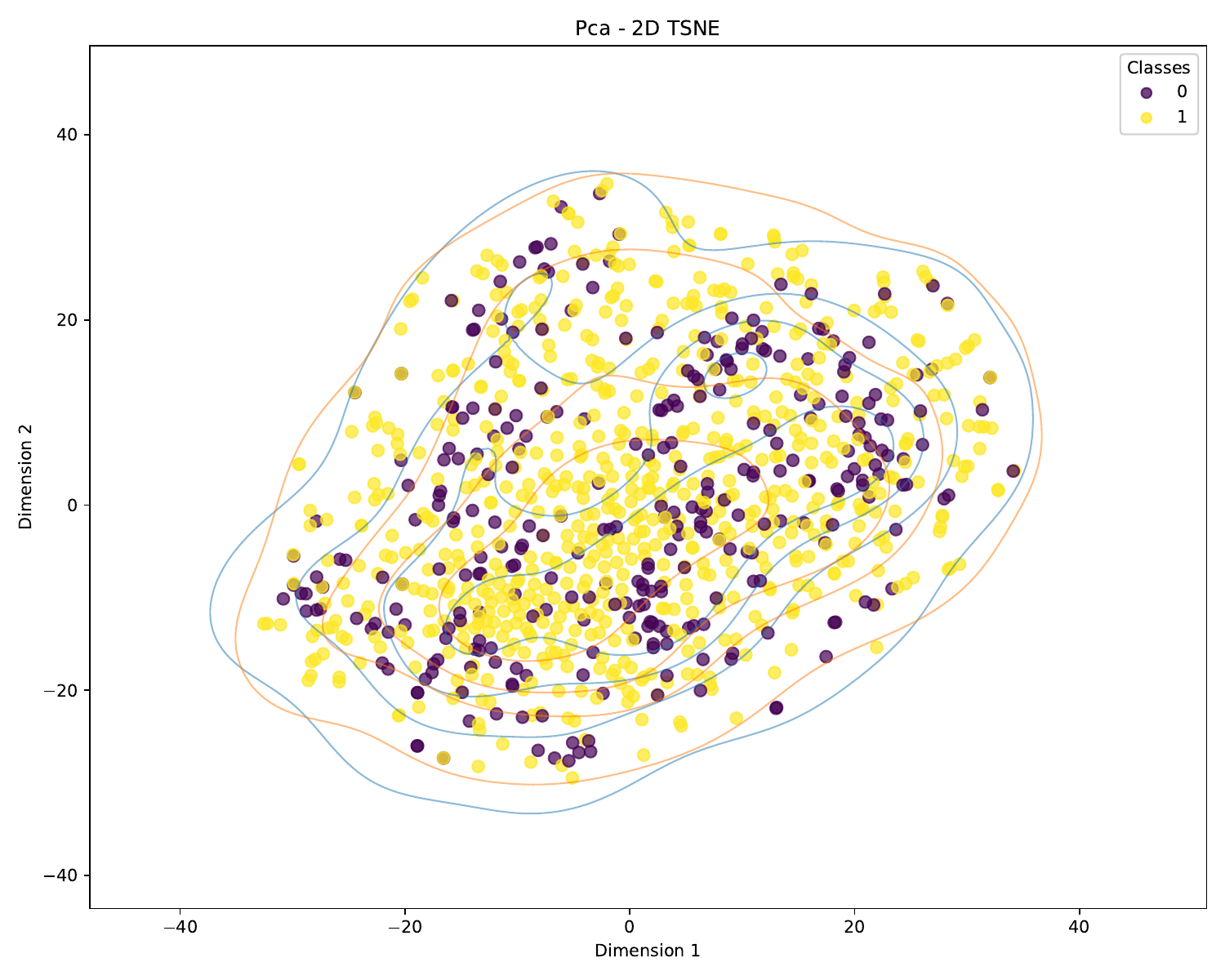}
  \includegraphics[width=0.32\textwidth]{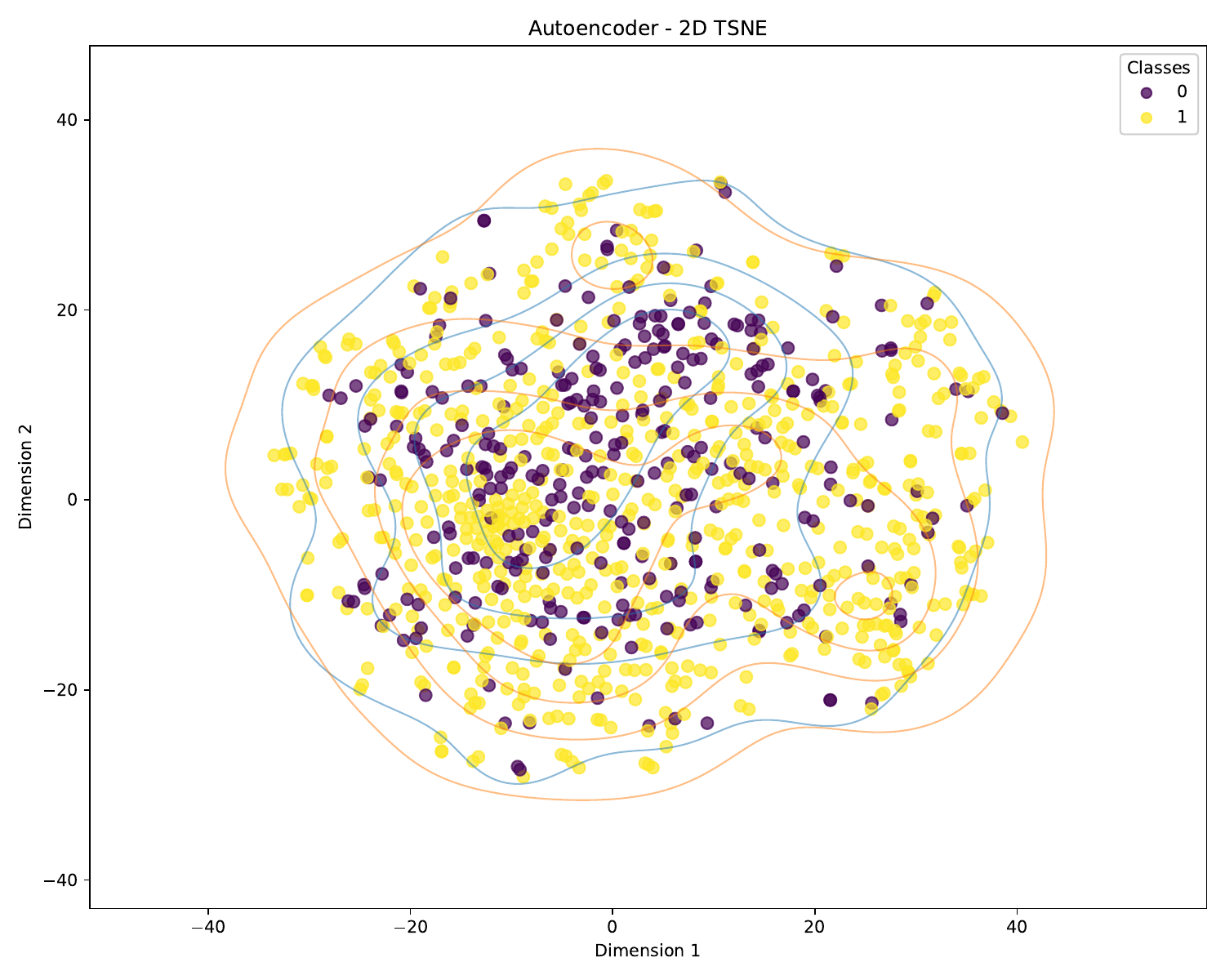}
  \includegraphics[width=0.32\textwidth]{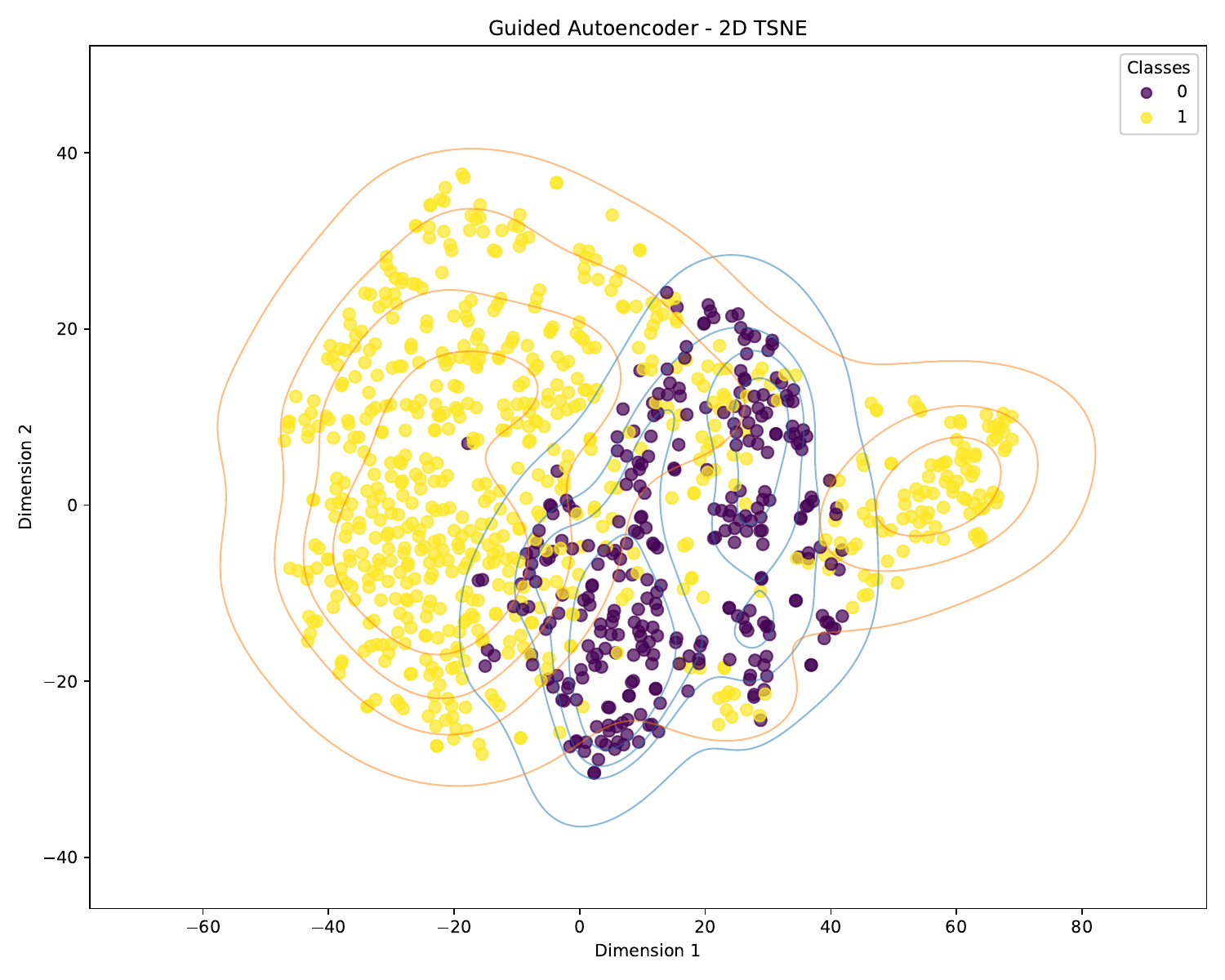}
  \caption{two-dimensional t-SNE projections of the 4768-dimensional QRC embeddings for (left) PCA + QRC, (center) AE + QRC, (right) QGARS + QRC}
  \label{fig:embeddings_tsne}
\end{figure*}

We also evaluate the effect of the number of qubits $N$ on the performance with the different dimensionality reduction methods as shown in Fig.~\ref{fig:n_qubits_accuracy} in general, it's easier to train the model as we increase $N$, the slight performance drop as we go from $N = 10$ to $N = 12$ might just be a byproduct due to the higher dimensionality of $\bm q$, we expect that with a larger amount of input images, the accuracy would keep improving. We hope to better understand this effect with future experiments.

In order to have a visual demonstration of how well the quantum reservoir embeddings separate classes, we project the high-dimensional embeddings down to three dimensions via t-distributed Stochastic Neighbour Embedding (t-SNE). Fig.~\ref{fig:embeddings_tsne} shows side-by-side plots for PCA → QRC, AE → QRC, and QGARS → QRC pipelines. The QGARS embeddings exhibit the tightest and most distinct clusters, indicating superior discriminative structure.

\subsection{Ablation Studies}

\begin{figure}[tb]
  \centering
  \includegraphics[width=0.8\linewidth]{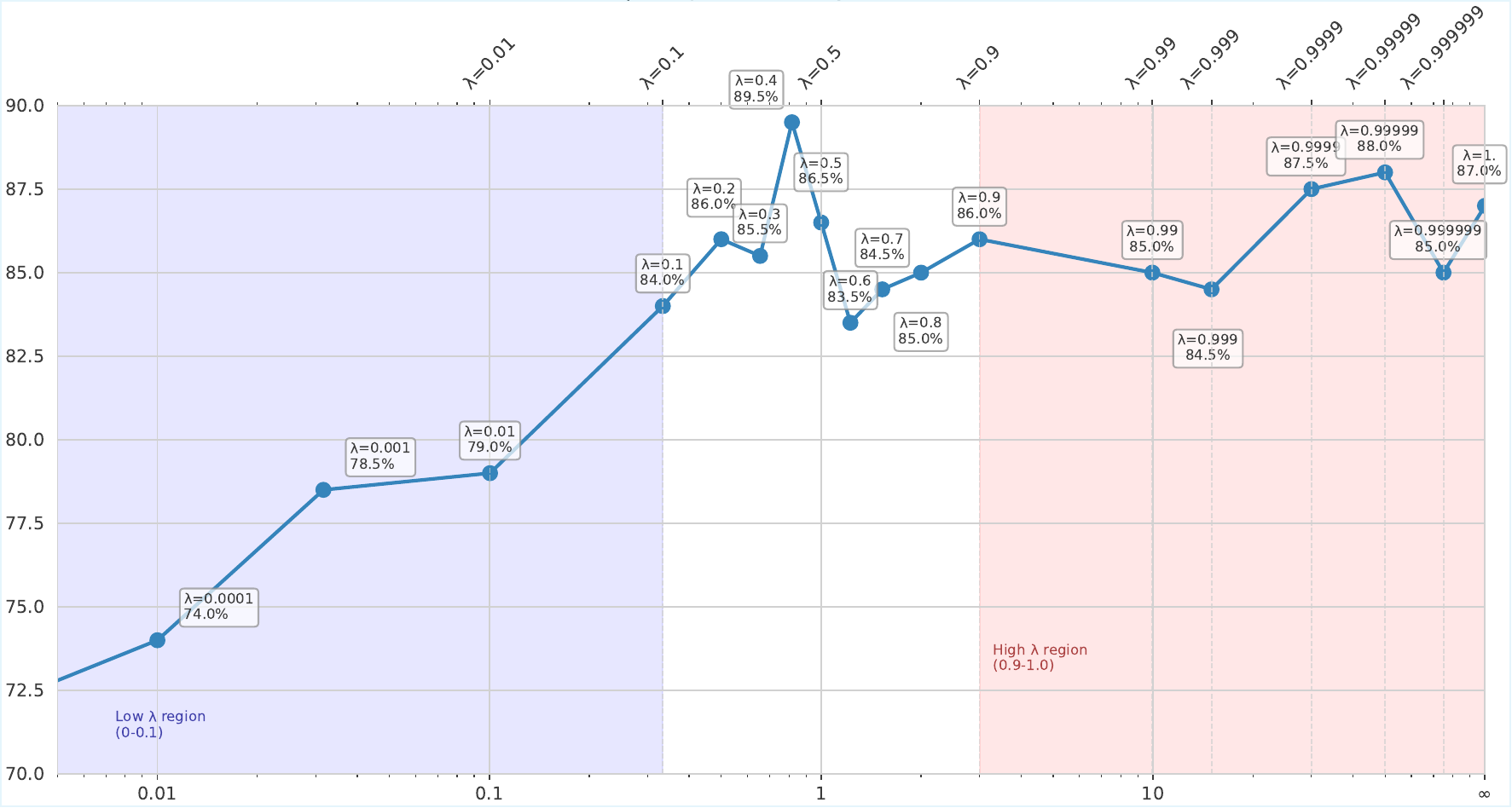}
  \caption{Accuracy as a function of the reconstruction/classification trade-off parameter $\lambda$.}
  \label{fig:lambda_sweep}
\end{figure}

We verified that varying the guidance weighting $\lambda$ in~\eqref{eq:composed_loss_function} has significant impact on the performance of our accuracy. To assess its influence, Fig.~\ref{fig:lambda_sweep} plots test accuracy vs.\ $\lambda$.

\noindent
We find optimal performance around $\lambda=0.5$; too little guidance ($\lambda<0.1$) approaches a point where it is equivalent to a traditional AE, so it yields under-trained classifiers. 

A strange effect that we verified was that even using the value \(\lambda = 1\), the performance of the classifier is still competitive. Although we still need to study this phenomenon, it suggests that the classical part of the autoencoder might not be as important as the surrogate component when producing a latent space that is appropriate for QRC classification.

In future work, we plan to to conduct more ablation studies, mainly on the effect that the frequency with which we update the surrogate model and the influence of the quantum parameters in the reservoir layer.




\subsection{Comparison with Classical Methods}
To benchmark the benefit of our QGARS pipeline, 
we compare against purely classical feature-reduction 
approaches followed by standard classifiers. 
Table~\ref{tab:classical_comp} reports test accuracy on each dataset 
for the different methods.

\begin{table}[tb]
  \caption{Classification accuracy (\%) for classical baselines (2000 training images and 400 testing images)}
  \label{tab:classical_comp}
  \centering
  \begin{tabular}{lccc}
    \hline
    Method            & Synthetic & CVC-ClinicDB & MNIST (0/1) \\
    \hline
    PCA + Linear      &  82.00 ± 0.88    &  71.05 ± 1.51   &  99.90 ± 0.14   \\
    AE + Linear       &  80.30 ± 3.37    &  68.25 ± 2.58   &  98.80 ± 0.33   \\
    PCA + NN          &  85.85 ± 1.39   &  91.65 ± 1.55   &  99.80 ± 0.21    \\
    AE + NN           &  83.75 ± 2.47    &  90.50 ± 1.93   &  99.90 ± 0.14    \\
    \hline
  \end{tabular}
\end{table}

\noindent
In contrast, the QGARS pipeline achieves a maximum of $\mathbf{92.0\%}$, $\mathbf{90.5\%}$ ans $\mathbf{99.0\%}$ on the same datasets (see Table~\ref{tab:qrc_reduction_accuracy}).

\subsubsection{Training Loss Dynamics}

\begin{figure}[!tb]
  \centering
  \includegraphics[width=0.9\linewidth]{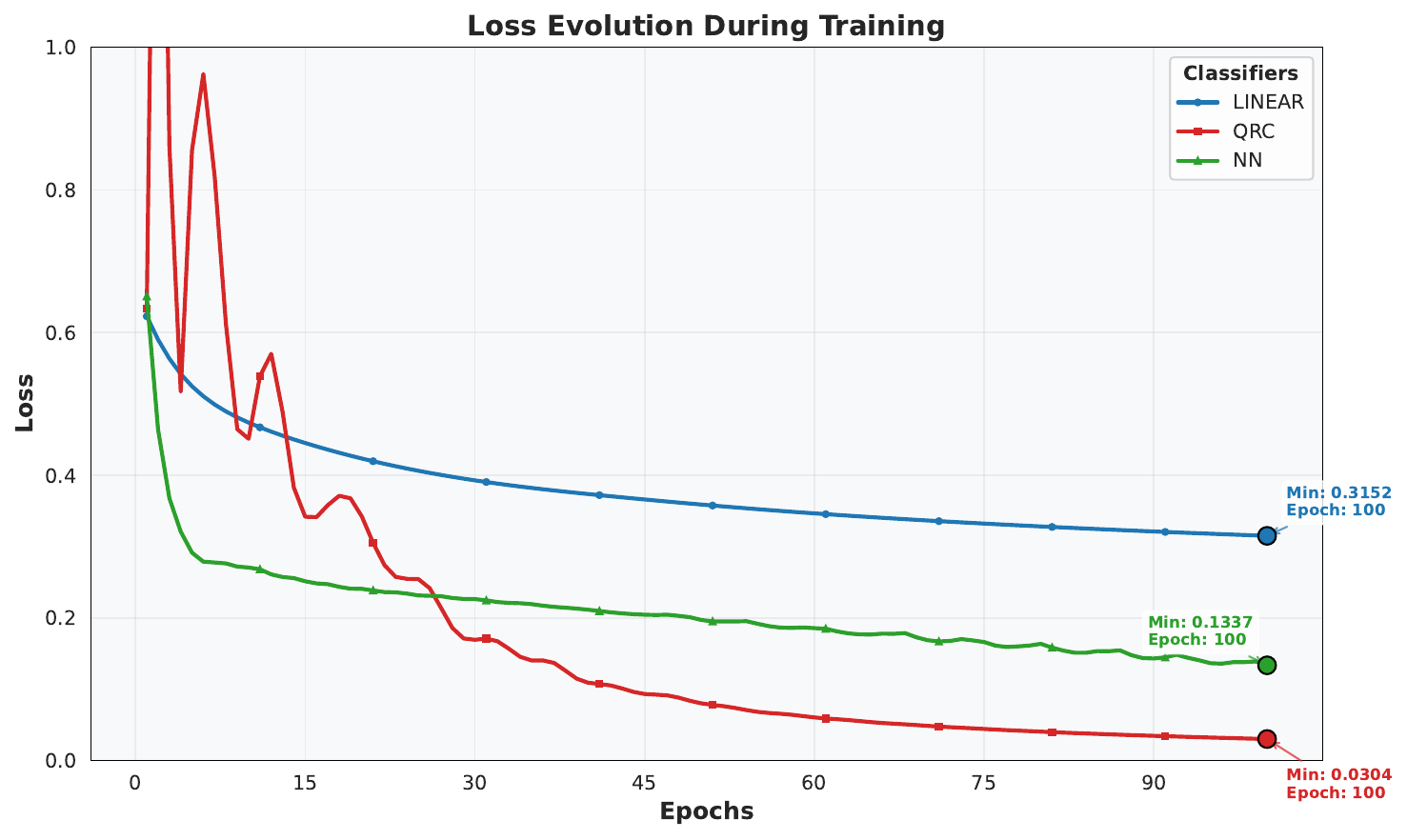}
  \caption{Training loss vs. epoch for PCA+Linear, AE+Linear, AE+NN, and QGARS with \( \lambda=0.7 \) on the Synthetic Polyp dataset.}
  \label{fig:loss_curves}
\end{figure}

Fig.~\ref{fig:loss_curves} shows the training loss curves for the classical baselines and QGARS on the Synthetic Polyp dataset. We plot the reconstruction loss for PCA/AE methods combined with classification loss for the AE+NN and QGARS pipelines.

\noindent
Although the loss function of the QGARS pipeline shows signs of instability, which can be attributed to the high dimensionality of the quantum embeddings, it reaches a lower loss in comparison to the classical counterparts.

\section{Limitations and Future Work}
\subsection{Current Limitations}
While QGARS demonstrates strong performance, several limitations remain:
\begin{itemize}
    \item \textbf{Classical overhead:}  
    Frequent surrogate retraining and batch quantum simulations introduce non-trivial classical compute overhead, potentially offsetting some of the quantum advantage.

    \item \textbf{Data and Task Specificity:}
    The current architecture has been evaluated primarily on specific medical imaging tasks. Its generalization to other domains or types of data remains to be thoroughly investigated.

\end{itemize}

\subsection{Potential Extensions}
Several avenues can further enhance QGARS:
\begin{itemize}
    \item \textbf{Different Hamiltonians:}
    Explore alternative interaction Hamiltonians (such as the one from the Heisenberg XXZ model) to better tailor reservoir dynamics, as experimented by A. D. Lorenzis et al.~\cite{lorenzisHarnessingQuantumExtreme2025}.

    \item \textbf{Convolutional Autoencoders:}
    Replace the fully connected autoencoder with a convolutional AE~\cite{lerchConvolutionalAutoencodersSpatiallyinformed2022} to learn spatially localized features. That would not only allow the model to better manage input size but also provide a richer structure for the reservoir.
    

    
    \item \textbf{Surrogate Model Complexity:} The surrogate model required to emulate the complex quantum reservoir has a very large number of parameters. We plan to investigate alternative architectures, such as Physics-Informed Neural Networks (PINNs)~\cite{raissiPhysicsInformedDeep2017}, which use physical laws to potentially improve model fidelity.

    \item \textbf{Experiment on Real Quantum Hardware:}
    Although the experiments we conducted on simulators showed promising results, to validate the usefulness of our pipeline in the NISQ era, we still need to run experiences on real quantum processing units (QPUs) with more qubits.

\end{itemize}

\section{Conclusion}

We have presented Quantum-Guided Autoencoder with Reservoir Surrogate (QGARS), a hybrid quantum-classical architecture that integrates a classical autoencoder with a neutral-atom quantum reservoir and a differentiable surrogate model to enable end‐to‐end training. By jointly optimizing reconstruction and classification losses, QGARS learns compact and discriminative features tailored for quantum reservoir computing. Our experiments on synthetic polyp, CVC-ClinicDB, and binary MNIST datasets demonstrate that QGARS outperforms classical PCA‐ and autoencoder‐based baselines that are commonly used for quantum reservoir computing pipelines. Ablation studies highlight the importance of guidance weight \(\lambda\), surrogate update frequency, and reservoir parameters. Although challenges remain—particularly regarding high‐dimensional embeddings, surrogate fidelity, and classical overhead, our proposed extensions and hardware considerations chart a clear path toward robust, scalable quantum‐enhanced medical imaging pipelines in the NISQ era.

\bibliographystyle{IEEEtran}
\bibliography{references}

\end{document}